%% file: main.tex
\documentclass[10pt,twocolumn,letterpaper]{article}

\usepackage{cvpr}              
\usepackage{multirow}
\usepackage{pifont}
\input{preamble}

\definecolor{cvprblue}{rgb}{0.21,0.49,0.74}
\usepackage[pagebackref,breaklinks,colorlinks,citecolor=cvprblue]{hyperref}

\title{
LRP-QViT: Mixed-Precision Vision Transformer Quantization via Layer-wise Relevance Propagation}

\author{Navin Ranjan \hspace{0.75cm} Andreas Savakis \\
Rochester Institute of Technology\\
Rochester, New York 14623, USA\\
{\tt\small nr4325@rit.edu} \hspace{0.5cm}{\tt\small andreas.savakis@rit.edu}}

\begin{document}
\maketitle
\input{sec/0_abstract}

\input{sec/1_intro}

\input{sec/2_related_work}
\input{sec/3_methodology}
\input{sec/4_Results}

{
    \small
    \bibliographystyle{ieeenat_fullname}
    \bibliography{main}
}


\end{document}

%% file: preamble.tex
\usepackage[dvipsnames]{xcolor}


%% file: sec/0_abstract.tex
\begin{abstract}

Vision transformers (ViTs) have demonstrated remarkable performance across various visual tasks. However, ViT models suffer from substantial computational and memory requirements, making it challenging to deploy them on resource-constrained platforms. Quantization is a popular approach for reducing model size, but most studies mainly focus on equal bit-width quantization for the entire network, resulting in sub-optimal solutions. While there are few works on mixed precision quantization (MPQ) for ViTs, they typically rely on search space-based methods or employ mixed precision arbitrarily. In this paper, we introduce LRP-QViT, an explainability-based method for assigning mixed-precision bit allocations to different layers based on their importance during classification. Specifically, to measure the contribution score of each layer in predicting the target class, we employ the Layer-wise Relevance Propagation (LRP) method. LRP assigns local relevance at the output layer and propagates it through all layers, distributing the relevance until it reaches the input layers. These relevance scores serve as indicators for computing the layer contribution score. 
Additionally, we have introduced a clipped channel-wise quantization aimed at eliminating outliers from post-LayerNorm activations to alleviate severe inter-channel variations. To validate and assess our approach, we employ LRP-QViT across ViT, DeiT, and Swin transformer models on various datasets. Our experimental findings demonstrate that both our fixed-bit and mixed-bit post-training quantization methods surpass existing models in the context of 4-bit and 6-bit quantization.


\end{abstract}

%% file: sec/1_intro.tex
\section{Introduction}
\label{sec:intro}

Vision Transformers (ViTs) have achieved state-of-the-art (SOTA) performance across various vision tasks, including image classification~\cite{image_16x16_dosovitskiy2021an_ViT, liu2021swin, wang2021PVT}, object detection~\cite{mao2022exploring_OD, carion2020end_end_OD} and segmentation~\cite{zheng2021rethinking_seg,strudel2021segmenter_seg}. However, their computational demands, high memory footprint, and significant energy consumption make them impractical for deployment on resource constrained platforms. 
Compression and acceleration techniques have been investigated for ViTs, aiming to reduce the original network size while maintaining  performance. Various approaches for model reduction include network pruning~\cite{he2018soft_prun, yu2022width_prun}, low-rank decomposition~\cite{denil2013predicting_lw}, quantization~\cite{lin2022fq_vit, li2023repq, li2022QViT}, knowledge distillation~\cite{chen2022dear_KD, lin2022knowledge_KD}, and dynamic token reduction~\cite{rao2021dynamicvit, xu2022evo_token}. 

Model quantization is an effective and popular approach to reduce the model size by converting floating-point parameters into lower-bit representations, but its drawback is an associated drop in performance. Thus, research focused on quantization methods that mitigate the performance drop is of significant importance. Quantization-Aware Training (QAT) involves quantizing model parameters to lower bit  precision and retraining on the entire dataset to optimize the quantized model and recover from the performance drop. 
However, this approach results in significant training costs in time and resources. In contrast, Post Training Quantization (PTQ) is considered an efficient and practical compression technique. It directly quantizes the model without the need for retraining. 
Instead, it simply uses a small sample of images to calibrate the quantized parameters.

Various studies have been conducted using PTQ for ViT quantization~\cite{lin2022fq_vit, ding2022APQViTtowards, yuan2022ptq4vit, li2023repq}. These works have identified several bottleneck components that limit the performance of the quantized model, such as LayerNorm, Softmax and GELU, and proposed PTQ schemes with improvements. However, most of the existing work is based on the premise that all the layers in ViTs contribute equally towards the model output and therefore use equal bit precision for the entire network leading to sub-optimal solutions. Fixed bit quantization forces important layers to be quantized with the same bit precision as the unimportant ones, missing an opportunity to further reduce the model size and enhance performance. Mixed-precision quantization (MPQ) addresses this limitation by allowing different bit precision for different layers. This enables the crucial layer to use higher bit precision than others. Most of the prior works for MPQ focus on convolutional neural networks (CNNs) and utilize policy search based methods~\cite{liu2021post_Rank_Aware_PTQ, lou2020autoq, xiao2023patch-wise}, or criterion based methods~\cite{dong2019hawq, dong2020hawqv2} to determine the optimal bit precision. 

In this paper, we propose an explainability approach for mixed-bit quantization of vision transformer models~\cite{LRP_chefer2021transformer, chefer2021LRP}, based on the contribution of each layer to the model performance. 
We utilize layer relevance propagation to obtain contribution scores for all layers that inform a mixed-precision bit allocation strategy for the quantization of different layers.

In our framework, we adopt and improve the RepQ-ViT~\cite{li2023repq} framework for model quantization. Specifically, for the post-LayerNorm activation, 
we introduce clipped channel-wise quantization to remove outliers and mitigate the effects of excessive inter-channel variation at the inference stages. This clipping is achieved by adjusting LayerNorm's affine factors and next layer weights. For post-Softmax activations, we fully adopt the modification made in~\cite{li2023repq}, which is initially quantized with $log\sqrt{2}$ quantizer to achieve higher representation for accuracy and rescale to $log2$ quantizer during the inference process for friendly quantization. With the improved quantization model and explainable mixed precision bit allocation, our main contributions are summarized as follows.
\begin{itemize}
    \item We propose LRP-QViT, a mixed precision framework for the quantization of vision transformer models.
    \item LPR-QViT uses layer relevance propagation to assign contribution scores to each layer and uses them to guide the bit allocation.
    \item We additionally introduce clipped channel-wise quantization for post-LayerNorm activation, that removes outliers and improves performance for both fixed-bit and mixed-precision quantization.
    \item Our results on ViT, DeiT and Swin  models demonstrate the superiority of LRP-QViT over fixed-bit quantization methods on classification, detection and segmentation benchmarks. 
\end{itemize}


%% file: sec/2_related_work.tex
\section{Related Work}
\label{sec:related_work}

\subsection{Vision Transformer}
The success of transformers in natural language processing (NLP) has inspired the application of vision transformer models on computer vision tasks. 
The seminal ViT architecture~\cite{image_16x16_dosovitskiy2021an_ViT} takes 
image patches as tokens and applies a transformer-based model for image classification, achieving remarkable success, which is largely attributed to the global receptive fields captured by the powerful self-attention (SA) mechanism. After the success of ViT, several works such as Swin~\cite{liu2021swin}, DeiT~\cite{touvron2021training_DeiT}  and PVT~\cite{wang2021PVT}, have further improved performance. DeiT~\cite{touvron2021training_DeiT} introduced a knowledge distillation strategy that relies on distillation tokens and employed various data augmentation techniques, significantly enhancing the effectiveness and efficiency of ViTs. Swin~\cite{liu2021swin} adopts a hierarchical architecture with shifted windows, capturing both local and global contextual information for improved performance. Beyond image classification, vision transformers have been successfully applied to other tasks, such as object detection~\cite{carion2020DETR,dai2021upDETR}, segmentation~\cite{jain2023oneformer}, and pose estimation~\cite{xu2022vitpose}. 

To address the high computational demand of ViTs, various efforts have been directed towards developing smaller and faster variants, such as DynamicViT~\cite{rao2021dynamicvit}, EfficientFormer~\cite{li2022efficientformer}, TinyViT~\cite{wu2022tinyvit} and MiniViT~\cite{zhang2022minivit}. However, these models retain full-precision parameters. In this paper, we focus on model quantization to address the challenge of high computational resources.

\subsection{Model Quantization}
Quantization reduces memory usage and computational demand by representing network parameters with lower precision than the standard full-precision model. Previous research has mainly focused on employing fixed-bit precision across all the layers in both QAT~\cite{li2022QViT,li2023Ivit} and PTQ~\cite{yuan2022ptq4vit, lin2022fq_vit,ding2022APQViTtowards,liu2023noisyquant,li2023repq}. 
PTQ stands out as an efficient and practical compression approach, as it directly quantizes the model without retraining. PTQ4ViT~\cite{yuan2022ptq4vit} used twin uniform quantization method to reduce quantization error in softmax and GELU activations and proposed a Hessian guided metric to search for the quantization scale. APQ-ViT~\cite{ding2022APQViTtowards} introduced a block-wise calibration scheme to address the optimization challenges in quantized networks and proposed Matthew-effect preserving quantization to maintain the power-law redistribution of the softmax layer. FQ-ViT~\cite{lin2022fq_vit} proposed the Power-of-Two Factor to handle inter-channel variation in LayerNorm and Log-Int-Softmax to quantize softmax layer. RepQ-ViT~\cite{li2023repq} separated the quantization and inference stages to achieve accurate quantization and efficient inference. In the quantization stage, it used channel-wise quantization to address inter-channel variation in LayerNorm and log$\sqrt{2}$ quantization to enhance the representation capabilities of softmax layer. During the inference stage, it reparameterized channel-wise to layer-wise quantization and log$\sqrt{2}$ to log2 quantization for hardware-friendly inference. 

Most existing works on mixed precision are primarily focused on CNNs using criterion based methods~\cite{dong2019hawq, dong2020hawqv2}, or search-based methods~\cite{wang2019haq, lou2020autoq}. There are few works on mixed-precision for ViTs~\cite{liu2021post, xiao2023patch-wise}. In~\cite{liu2021post}, mixed precision for the multi-head self-attention (MHSA) and multi-layer perceptron (MLP) module is based on sensitivity, which is calculated using the nuclear norm of the attention map for MHSA and the output feature for MLP. PMQ~\cite{xiao2023patch-wise} estimated the sensitivity of the layer by measuring the error induced by the network when the layer is removed and used a Pareto frontier approach to allocate optimal bit widths. In this paper, we leverage transformer explainability to estimate the ViT layer sensitivity for allocating mixed precision for quantization. Furthermore, we adopt and improve the RepQ~\cite{li2023repq} framework for efficient mixed precision PTQ.

\subsection{Explainability for Transformers}
The majority of prior research on explainability has concentrated on CNNs, with a main focus on gradient methods~\cite{srinivas2019fullgrad, smilkov2017smoothgrad} or attribution methods~\cite{binder2016layerLRP_NN, gu2019understanding}. However, as vision transformer architectures have advanced, there has been a notable upswing in research on explainability for transformers. In the Attention Rollout method~\cite{abnar2020quantifying_attentionrollout}, attention scores from different layers are linearly combined, yet this approach struggles to distinguish between positive and negative contributions.
The Layer-wise Relevance Propagation method propagates relevance from the predicted class backward to the input image. Several works have applied LRP to Transformers~\cite{voita2019analyzing, binder2016layerLRP_NN}. However, many of these studies overlook the propagation of attention across all layers and neglect parts of the network that perform mixing of two activation maps, such as skip connections and matrix multiplication. Additionally, most of the work does distinguish between the positive and negative contribution provided by the layers toward the model decision. Without such distinction, both positive and negative contributions are mixed, resulting in higher than necessary relevance scores. 
 
In~\cite{chefer2021LRP}, relevance and gradients information are used in a way that iteratively removes the negative contribution and calculate the accurate relevance score for each attention head in each layers of a transformer model. In this work, we adopt the relevance score calculation mentioned in ~\cite{chefer2021LRP} and extend the approach to calculate the relevancy score for other layers, such as qkv layer, linear projection layer, matrix multiplication layers, and fully connected layers.

%% file: sec/3_methodology.tex
\section{Methodology}
\label{sec:methodology}

\subsection{Vision Transformer Architecture}
The vision transformer takes an image and reshapes into a sequence of $N$ flattened 2D patches. Each patch is mapped into a vector of $D$ dimensions via a linear projection layer, denoted by $X\in{R}^{N\times D}$. The core structure of the standard ViT contains several blocks, each consisting of a MHSA module and a MLP module. The MHSA module is designed to understand the relationship between tokens, extracting features with a global perspective. In the $l^{th}$ transformer block, the MHSA module process the input sequence $X^{l}$, which undergoes initial linear projections to obtain the query $Q_{l} = X^{l}W^q_{i}$, key $K_{l} = X^{l}W^k_{i}$, and value $V_{l} = X^{l}W^v_{i}$, abbreviated as qkv. Subsequently, attention scores are computed between queries and keys, followed by a softmax layer. The final output of MHSA is obtained by concatenating the outputs from multiple heads within the MHSA, as follows: 
\begin{align}
\text{Attn}_{i} = \text{Softmax}\left(\frac{Q_{i}\cdot K_{i}^T}{\sqrt{D_{h}}}\right)V_{i},
\label{eq:1} \\
\text{MHSA}(X^{l}) = [\ \text{Attn}_{1}, \text{Attn}_{2},...,\text{Attn}_{i}]\ W^o,
\label{eq:2}
\end{align}
where $h$ is the number of attention heads and $D_{h}$ is the feature size of each head, and $i = 1, 2, ..., h$. The MLP module employs two fully connected layers separated by a GELU activation to project the features into a high-dimensional space and learn representations, as follows:
\begin{align}
\text{MLP}(Y^{l}) = \text{GELU}\left(Y^{l}W^{1} + b^{1} \right)W^{2} + b^{2}.
\label{eq:3}
\end{align}
where $Y^{l}$ denotes input for MLP, $W^{1}\in \mathbb{R}^{D \times D}$, $b^{1}\in\mathbb{R}^{D_f}$, $W^{2}\in \mathbb{R}^{D_{f} \times D}$, and $b^{2}\in\mathbb{R}^{D}$. The LayerNorm $(LN)$ are applied before each modules and residuals are added after each module, the transformer block is formulated as:
\begin{align}
Y^{l} = X^{l} + \text{MHSA}\left(LN\left(X^{l}\right)\right),
\label{eq:4} \\
X^{l+1} = Y^{l} + \text{MLP}\left(LN\left(Y^{l}\right)\right).
\label{eq:5}
\end{align}
The large matrix multiplications in MHSA and MLP contribute significantly to computational costs. Following 
~\cite{liu2021post, li2023repq}, we quantize all the weights and inputs involved in matrix multiplication, linear embedding, and softmax while keeping layer normalization at full precision. 
\subsection{Model Quantization}
In this paper, we quantize the weights and activations of linear layers, convolutional layers, and matrix multiplication using uniform quantizer function and quantized softmax activation using a $log2$ quantizer function. The uniform quantization splits the data range equally and is defined as: 
\begin{align}
\text{Quant: } x^{q} &= \text{clip}\left(\left\lfloor\frac{x}{s}\right\rceil + z, 0, 2^{b}-1\right) \label{eq:6} \\
\text{DeQuant: } \bar{x} &= s \cdot (x^{q} - z) \approx x \label{eq:7} \\
s &= \frac{\max(x) - \min(x)}{2^b-1}, \quad \text{and} \notag \\
z &= \left\lfloor -\frac{\min(x)}{s} \right\rceil \label{eq:8}
\end{align}
Here, $x$ is the original floating-point weights or inputs, $x^{q}$ represents quantized values, b is the quantization bit-precision,  $\lfloor\cdot\rceil$ denotes the round operator, and $\text{clip}$ denotes the elements in the tensor that exceed the ranges of the quantization domain are clipped.
$s$ is the quantization scale and $z$ is the zero-point, both of which is determined by the lower and upper bound of $x$. In de-quantization process, the de-quantized value $\bar{x}$ approximately recovers $x$.

The $log2$ quantization function converts the quantization process from linear to exponential and is defined as:
\begin{align}
\text{Quant: } x^{q} &= \text{clip}\left(\left\lfloor -\log_2\frac{x}{s} \right\rceil, 0, 2^{b}-1\right) \label{eq:9} \\
\text{DeQuant: } \bar{x} &= s \cdot 2^{-x^{q}} \approx x
\label{eq:10} 
\end{align}

\subsubsection{Clipped Reparameterization for LayerNorm Activations}
In ViTs, during inference, LayerNorm computes the statistics $\mu_x$, and $\sigma_x$ in each forward step and normalizes the input $X\in\mathbb{R}^{N\times D}$. Then, affine parameters $\gamma\in\mathbb{R}^D$ and $\beta\in\mathbb{R}^D$ re-scale the normalized input to another learned distribution. The LayerNorm process is defined as:
\begin{align}
\text{LayerNorm(X)}=\frac{X-\mu_x}{\sqrt{{\sigma_x^{2}}+\epsilon}}\odot \gamma + \beta,
\label{eq:11}
\end{align}
where $\odot$ denotes Hadamard product.

Looking into the Post-LayerNorm activations, we observe an extreme inter-channel variation, a critical factor that reduces post training quantization performance. Several studies have investigated this issue and proposed solutions. In~\cite{shen2020qQBert}, group-wise quantization was implemented, treating individual matrices associated with a head in MHSA as one group and assigning different quantization parameters to each group. In~\cite{lin2022fq_vit}, channel-wise quantization was applied, providing different channels with different parameters based on a power-of-two-factor approach. In~\cite{li2023repq}, quantization-inference decoupling techniques were used, performing quantization using a channel-wise approach. Then, during inference, channel-wise quantized parameters were re-parameterized to layer-wise quantization by taking an average. 

Taking inspiration from~\cite{lin2022fq_vit, li2023repq}, we adopt the quantization-inference decoupling paradigm and propose a simple but effective strategy using Clipped Reparameterization for LayerNorm activations (CRL). This involves restricting inter-channel variations by clipping outliers within a set number of standard deviations around the channel mean value. Specifically, for $l^{th}$ transformer block, given the input $X^{l}_{LN}$, we perform channel-wise quantization to obtain the scale $s\in\mathbb{R}^D$ and zero-point $z\in\mathbb{R}^D$. Next, we calculate a clipped channel-wise quantized scale $\hat{s}$ and zero-point $\hat{z}$, as follows:
\begin{align}
\hat{s} = \text{clip}(s, \mu_s-2\sigma, \mu_s+2\sigma) \notag \\
\hat{z} = \text{clip}(z, \mu_z-2\sigma, \mu_z+2\sigma)
\label{eq:12} 
\end{align}
The variation factors between the original and clipped parameters are denoted as $v_1=s/\hat{s}$ and $v_2=z-\hat{z}$. Eq. (\ref{eq:8}) can then be expressed as:
\begin{align}
\hat{s} = \frac{s}{v_1} = \frac{\left[\max(x) - \min(x)\right]/v_1}{2^b - 1} \label{eq:13} \\
\hat{z} = z - v_2 = \left\lfloor -\frac{\min(x) + s \odot v_2}{s} \right\rceil \label{eq:14}
\end{align}
In Eq. (\ref{eq:13}), dividing each channel of $X^l_{LN}$ by $v_1$ results in $\hat{s}$. Similarly, in Eq. (\ref{eq:14}), adding $s\odot v_2$ to each channel of $X^l_{LN}$ gives $\hat{z}$. These operations can be achieved by adjusting the LayerNorm's affine factors as follows:
\begin{align}
\hat{\beta} = \frac{\beta+s\odot v_2}{v_1},
\text{\hspace{0.1in}}
\hat{\gamma} =\frac{\gamma}{v_1} 
\label{eq:15}
\end{align}
This reparameterization induces a change in the activation distribution, specifically expressed as $\hat{X}^l_{LN} = \left(X^l+s\odot v_2\right)/v_1$. 
In the MHSA module, the layer following LayerNorm is a linear projector layer for qkv. The reparameterized shift in the qkv layer is expressed as
\begin{align}
X^l_{LN}\cdot W^{qkv} = \frac{X^l_{LN}+s\odot v_2}{v_1}\left(v_1\odot W^{qkv}\right) \notag \\+ \left(b^{qkv}-\left(s\odot v_2\right)W^{qkv}\right)
\label{eq:16}
\end{align}
Here, $W^{qkv}\in \mathbb{R}^{D\times D_{h}}$ and $b^{qkv} \in \mathbb{R}^{3D_h}$ represent the weight and bias of the qkv layer. To offset this distribution shift, the weight of the subsequent layer can be adjusted, as outlined below:
\begin{align}
\hat{W}^{qkv} = v_1\odot W^{qkv} \notag \text{\hspace{0.5in}}
\\ 
\hat{b}^{qkv} = b^{qkv}-\left(s\odot v_2\right)W^{qkv}
\label{eq:17}
\end{align}
A similar adjustment strategy is applied to the input $Y^l$ in the MLP module. Consequently, through the modification of LayerNorm affine factors and the weights and bias of the next layer, we reparameterize channel-wise quantization to achieve clipped channel-wise quantization for LayerNorm activation, effectively addressing the issue of inter-channel variation.

\subsubsection{Nonlinear Quantization for Softmax Activations}

In ViTs, the softmax operation transforms the attention scores of the MHSA module into probabilities, exhibiting a power-law distribution that is highly unbalanced and unsuitable for quantization. We observe that the majority of the distribution comprises very small values, and only a few values have larger magnitudes. Previous approaches~\cite{lin2022fq_vit} directly applied a $log2$ quantizer, while in~\cite{ding2022APQViTtowards}, a Matthew-effect preserving quantization was proposed. Although these methods outperform uniform quantization, they do not consistently achieve satisfactory performance.

In~\cite{li2023repq}, a quantization-inference decoupling approach is employed. A $log\sqrt{2}$ quantizer is used to quantize the softmax activation, instead of $log2$, as it offers higher quantization resolution and accurately describes the power-law distribution. During the inference stage, the $log\sqrt{2}$ quantizer parameter is reparameterized to a hardware-friendly $log2$ quantizer, achieving both accuracy through $log\sqrt{2}$ and efficiency through $log2$. 
In this we adopt RepQ-ViT.

\subsection{Layer-Wise Relevance Propagation}
Layer-wise Relevance Propagation is an explainable artificial intelligence approach that provides insight into the contribution of each feature to the model output. It works by assigning a relevance score to the output layer and then back-propagating it through all layers of the network to the input features. These relevance scores serve as indicators for computing a contribution score for each layer. We use these contribution scores to assign mixed-precision bit allocation for PTQ.
 

\subsubsection{Relevance and Gradients}
Let $C$ be the number of classes of the classification head, and $c\in 1 \dots|C|$ the class to be visualized. We propagate relevance $\left(R\right)$ and gradients $\left(\nabla\right)$ with respect to class $c$. Let $x^{\left(n\right)}$ be the input of $L^{n}$ layer of of the network, where $n\in \left[1 \dots N\right]$, $x^{N}$ is the input and $x^1$ is the output of network. The gradients with respect to the classifier's output $y$, at class $t$, namely $y_t$, is given by 
\begin{align}
\nabla{x}^{n}_{j} = \frac{\partial{y}_t}{\partial{x}^{n}_{j}} = \sum_{i}{\frac{\partial{y}_t}{\partial{x}^{n-1}_{i}}\frac{\partial{x}^{n-1}_{i}}{\partial{x}^{n}_{j}}}
\label{eq:18}
\end{align}
where the index $j$ corresponds to elements in $x^n$ and $i$ corresponds to elements in $x^{n-1}$. Let $L^n_i\left(X,W\right)$ be layer operation on two tensor $X$ and $W$. Relevance propagation follows the generic Deep Taylor Decomposition~\cite{montavon2017explainingDTD} as follows:
\begin{align}
R^{n}_{j} = \mathcal{G}\left(X,W, R^{n-1}\right) \notag 
{\textbf{\hspace{1.2in}}} \\
= \sum_{i}X_j\frac{\partial L^{n}_j\left(X,W\right)}{\partial X_j}\frac{R^{n-1}_i}{\sum_{j'}L^n_{j'}\left(X,W\right)}
\label{eq:19}
\end{align}
The transformer block consist of GELU~\cite{hendrycks2016GELU}, which outputs both positive and negative values. To preserve the conservation rule, i.e., sum of relevance is always same for all the layers, we remove all the elements with negative relevance. We construct a subset of indices $p = \{\left(i,j\right)|x_jw_{ji}\geq )\}$, resulting in the following propagation: 
\begin{align}
R^{n}_{j} = \mathcal{G}_{p}\left(x,w,p,R^{n-1}\right) \notag 
{\textbf{\hspace{1.0in}}} \\
= \sum_{\left(i,j\right)\in p}\frac{x_jw_{ji}}{\sum_{\{j'|\left(j',i\right)\in p\}}x_{j'}w_{j'i}}R^{n-1}_i
\label{eq:20}
\end{align}
To initiate the relevance propagation, we set $R^0=1_c$, where $1_c$ is a one-hot indicating the target class $c$.

\subsubsection{Layer Importance Score}
In LRP, both the relevance and gradients propagate backward from the classification head through all layers within each block to the input patch embedding layers. During this process, each layer in the network learns a relevance score map. Specifically, for image sample $t$, for the attention layer $A$ in transformer block $L$, with the layer gradients denoted as $\nabla A^L_t$ and relevance as $R^L_{A,t}$, the relevance score map of the attention layer $\left(A^L_{s,t}\right)$ is defined as:
\begin{align}
{A}^{L}_{s,t} = \mathbb{E}_h\left(\nabla A^L_t \odot R^L_{A,t}\right)^+
\label{eq:21}
\end{align}
where $\mathbb{E}_h$ is the mean across the multi-head, and our analysis focuses exclusively on the positive value of the gradients-relevance multiplication. To quantitatively measure the contribution score of the attention layer of $L^{th}$ toward the output for the sample image, we simply take the mean of the relevance score map. Since the relevance score map produced by the LRP method is class-specific, i.e., different maps are generated for various image samples. The overall contribution score of the attention layer of the $L^{th}$ block is calculated by taking average over $T$=50,000 randomly selected images from the ImageNet1k training dataset, given as:
\begin{align}
{C}^{L}_A = \frac{1}{T}\sum_{t=1}^{T}\mathbb{E}\left({A}^{L}_{s,t}\right)
\label{eq:22}
\end{align}
The identical method is utilized for all other quantizing layers, including qkv layers, matrix multiplication layers (matmul1 and matmul2), the projection layer, and the fully connected layers (fc1 and fc2) to calculate their contribution score toward output classification. The relative importance score for a any layer in a any block is the average normalized value of the contribution score of that layer across all contribution scores. The expression for the relative layer importance score of attention layer in the $L^{th}$ block is as follows:
\begin{align}
{I}^{L}_A = \frac{C^L_A}{\sum_{l=1}^{L}\sum_{u \in U_l}C^l_u}
\label{eq:23}
\end{align}
Here, $U_l$ represents all the quantized layers in the $L^{th}$ transformer block. We use the relative importance score $I$ to determine the bit-width during model quantization, allocating a higher bit count to layers with higher relative importance scores.

\subsubsection{Mixed-Precision Bit Allocation}

The early blocks of the transformer architecture are typically responsible for capturing low-level features and details in the input data. These features are crucial for the network to learn representations effectively. In addition, early blocks are sensitive to small variations in the input data. If these blocks are quantized too aggressively, the network might lose its representation capacity, potentially leading to a decrease in its overall performance. Therefore, to avoid loss of information early in the processing stages, we allocate higher precision to all the layers of the first two blocks during post training quantization. To maintain the model size, we use the layer importance score to identify the layers that are important and reduce the bit allocation of those layers. 

%% file: sec/4_Results.tex
\section{Experiments}
\label{sec:result}

\subsection{Experimental Setup}
For post training quantization, all the pretrained weights and backbone architecture are adopted from Timms library. We randomly sample 32 images from the ImageNet1K training set for image classification and 1 sample from the COCO dataset for object detection and instance segmentation to calibrate the quantization parameters. We apply percentile method for the calibration process. Clipped scale reparameterization is applied to the post-LayerNorm activations in all the blocks and the scale reparameterization is applied to all the MHSA modules. For layer importance analysis, all the pretrained weights are obtained from Timms library and the backbone architecture is adopted from~\cite{chefer2021LRP}. 


\subsection{Results on ImageNet1K}
To demonstrate the superiority of LRP-QVIT in image classification, we conducted extensive experiments on the ImageNet1K~\cite{krizhevsky2012imagenet} dataset by employing various vision transformer architectures, including ViT~\cite{image_16x16_dosovitskiy2021an_ViT}, DeiT~\cite{touvron2021training_DeiT}, and Swin~\cite{liu2021swin}. The comparison of quantization results with PTQ methods is presented in Table~\ref{tab:ResultsImageNet}. Our fixed-bit clipped Reparameterized Layer method (CRL-QViT) and our mixed-bit method (LRP-QViT) outperform SOTA methods by significant margins. Specifically, for 4-bit weight and activation quantization, our fixed-bit CRL-QViT method achieves approximately (2-3)\% higher accuracy compared to the current state-of-the-art RepQ-ViT~\cite{li2023repq}. 
With the addition of the mixed-precision strategy, LRP-QViT significantly improves accuracy, averaging up to 4\% compared to RepQ-ViT. 

For specific models like ViT-S and ViT-B, our mixed-precision LRP-QViT outperforms RepQ-ViT by 5.76\% and 6.89\%, respectively. Similarly, for 6-bit quantization, fixed-bit CRL-QViT and mixed-bit LRP-QViT outperform all other methods. Our methods can achieve an accuracy comparable to that of the full-precision baseline. In DeiT-B and Swin-S quantization, mixed-precision LRP-QViT achieves 81.44\% and 82.86\% accuracy, respectively, with only a 0.36\% and 0.37\% accuracy drop over the full precision models. We also observe FQViT crashing when quantizing the model to 4 bits, while our methods achieve top performance. Furthermore, our methods do not depend on hyperparameter tuning or a reconstruction process, as seen in FQViT, PTQ4ViT, and APQ-ViT.

\begin{table*}
 \small
  \centering
  \begin{tabular}{lcccccccccr}
    \toprule
    Method&No HP&No REC.&Prec.(W/A)&ViT-S&ViT-B&DeiT-T&DeiT-S&DeiT-B&Swin-S&Swin-B\\
    \midrule
    Full-Precision&-&-&32/32&81.39&84.54&72.21&79.85&81.80&83.23&85.27 \\
    \hline
    FQ-ViT~\cite{lin2022fq_vit}&\ding{53} & \checkmark{}&4/4&0.10&0.10&0.10&0.10&0.10&0.10&0.10 \\
    PTQ4ViT-ViT~\cite{yuan2022ptq4vit}&\ding{53} & \ding{53}&4/4&42.57&30.69&36.96&34.08&64.39&76.09&74.02 \\
    APQ-ViT~\cite{ding2022APQViTtowards}&\ding{53} & \ding{53}&4/4&47.95&41.41&47.94&43.55&67.48&77.15&76.48 \\
    RepQ-ViT~\cite{li2023repq}&\checkmark&\checkmark&4/4&65.05&68.48&57.43&69.03&75.61&79.45&78.32\\
    CRL-QViT(ours)&\checkmark&\checkmark&4/4&68.25&73.54&59.06&70.78&77.40&80.55&80.04 \\
    LRP-QViT(ours)&\checkmark&\checkmark&MP4/MP4&\textbf{70.81}&\textbf{75.37}&\textbf{61.24}&\textbf{72.43}&\textbf{78.13}&\textbf{81.37}&\textbf{80.77} \\
    \hline
    FQ-ViT~\cite{lin2022fq_vit}&\ding{53} & \checkmark{}&6/6&4.26&0.10&58.66&45.51&64.63&66.50&52.09 \\
    PSAQ-ViT~\cite{li2022patch}&\ding{53} & \checkmark{}&6/6&37.19&41.52&57.58&63.61&67.95&72.86&76.44 \\
    Ranking~\cite{liu2021post}&\ding{53} & \ding{53}&6/6&-&75.26&-&74.58&77.02&-&- \\
    PTQ4ViT~\cite{yuan2022ptq4vit}&\ding{53} & \ding{53}&6/6&78.63&81.65&69.68&76.28&80.25&82.38&84.01 \\
    APQ-ViT~\cite{ding2022APQViTtowards}&\ding{53} & \ding{53}&6/6&79.10&82.21&70.49&77.76&80.42&82.67&84.18 \\
    RepQ-ViT~\cite{li2023repq}&\checkmark&\checkmark&6/6&80.43&83.62&70.76&78.90&81.27&82.79&84.57 \\
    CRL-QViT (ours)&\checkmark&\checkmark&6/6&80.52&83.83&70.96&79.00&81.39&82.77&84.63\\
    LRP-QViT (ours)&\checkmark&\checkmark&MP6/MP6&\textbf{80.59}&\textbf{83.87}&\textbf{71.03}&\textbf{79.03}&\textbf{81.44}&\textbf{82.86}&\textbf{84.72} \\
    \bottomrule
  \end{tabular}
  \caption{Quantization results of image classification on ImageNet1K dataset, each value presents the Top-1 accuracy (\%) obtained by quantizing each model.
  CRL-QViT utilizes Clipped Reparameterization for LayerNorm, while LRP-QViT incorporates both CRL and Layer-wise Relevance Propagation.
  Here, "No Hyper-parameters" is abbreviated as "No HP", "No Reconstruction" as "No REC", and "Prec. (W/A)" indicates the quantization bit-precision for weights and activations as "W" and "A" bits, respectively. 'MP' represents mixed precision. Bold data indicate best performance}
  \label{tab:ResultsImageNet}
\end{table*}

\subsection{Results on COCO }
To further assess the effectiveness of LRP-QViT, we conduct evaluations on object detection and instance segmentation tasks using the COCO~\cite{lin2014microsoft_COCO} dataset. Employing Mask R-CNN and Cascade Mask R-CNN detectors with Swin transformers as the backbone, we present the results in Table~\ref{tab:ResultsCOCO}. PTQ4ViT exhibits severe performance degradation, and APQ-ViT demonstrates improved results but still performs suboptimally with Swin-T. Moreover, both methods require hyperparameter tuning and a reconstruction process. In comparison to the state-of-the-art RepQ-ViT, our LRP-QViT method achieves superior performance. For 4-bit quantization using the Swin backbone in the Mask R-CNN framework, our mixed-precision LRP-QViT outperforms RepQ-ViT by an average of 2 box AP and 1.55 mask AP. In the Cascade Mask R-CNN framework, our mixed-precision LRP-QViT approach exhibits slightly better performance compared to RepQ-ViT. Similarly, for 6-bit quantization, LRP-QViT shows slightly better performance compared to RepQViT. When quantizing the Mask R-CNN framework with Swin-T, our approach achieves 45.6 box AP and 41.3 mask AP, which is only 0.4 box AP and 0.3 mask AP lower than the full-precision baseline. Similarly, comparable results can be obtained with the Swin-S backbone, achieving 48.1 box AP and 43.3 mask AP, which is just 0.4 box AP and 0.3 mask AP lower compared to full precision.

\begin{table*}
  \small
  \centering
  \scalebox{0.9}{
  \begin{tabular}{cccccccccccc}
    \toprule
    \multirow{3}{*}{Method} & \multirow{3}{*}{No HP} & \multirow{3}{*}{No REC} & \multirow{3}{*}{Prec. (W/A)} & \multicolumn{4}{c}{Mask R-CNN} & \multicolumn{4}{c}{Cascade Mask R-CNN} \\
    \cline{5-8} \cline{9-12}
    & & & & \multicolumn{2}{c}{w.Swin-T} & \multicolumn{2}{c}{w.Swin-S} & \multicolumn{2}{c}{w.Swin-T} & \multicolumn{2}{c}{w.Swin-S} \\
    & & & & AP$^{box}$ & AP$^{mask}$ & AP$^{box}$ & AP$^{mask}$ & AP$^{box}$ & AP$^{mask}$ & AP$^{box}$ & AP$^{mask}$ \\
    \midrule
    Full-Precision & - & - & 32/32 & 46.0 & 41.6 & 48.5 & 43.3 & 50.4 & 43.7 & 51.9 & 45.0 \\
    \hline
    PTQ4ViT~\cite{yuan2022ptq4vit} & \ding{53} & \ding{53} & 4/4 & 6.9 & 7.0 & 26.7 & 26.6 & 14.7 & 13.5 & 0.5 & 0.5 \\
    APQ-ViT~\cite{ding2022APQViTtowards} & \ding{53} & \ding{53} & 4/4 & 23.7 & 22.6 & 44.7 & 40.1 & 27.2 & 24.4 & 47.7 & 41.1 \\
    RepQ-ViT~\cite{li2023repq} & \checkmark & \checkmark & 4/4 & 36.1 & 36.0 & 44.2 & 40.2 & 47.0 & 41.4 & 49.3 & 43.1 \\
    CRL-QViT (ours) & \checkmark & \checkmark & 4/4 & 37.2& 37.4& 45.8 & 40.9 & 47.4 & 41.5 & 49.5 & 43.4  \\
    LRP-QViT (ours) & \checkmark & \checkmark & MP4/MP4 & \textbf{37.9} & \textbf{38.2} & \textbf{46.4} & \textbf{41.4} & \textbf{47.7} & \textbf{41.6} & \textbf{50.1} & \textbf{43.6} \\
    \hline
    PTQ4ViT~\cite{yuan2022ptq4vit} & \ding{53} & \ding{53} & 6/6 & 5.8 & 6.8 & 6.5 & 6.6 & 14.7 & 13.6 & 12.5 & 10.8 \\
    APQ-ViT~\cite{ding2022APQViTtowards} & \ding{53} & \ding{53} & 6/6 & 45.4 &41.2  &47.9  &42.9  &48.6  &42.5  &50.5  &43.9  \\
    RepQ-ViT~\cite{li2023repq} & \checkmark & \checkmark & 6/6 & 45.1 & 41.2 & 47.8 & \textbf{43.0} & \textbf{50.0} & \textbf{43.5} & 51.4 & \textbf{44.6} \\
    CRL-QViT (ours) & \checkmark & \checkmark & 6/6 & 45.3 & 41.2 & 47.9 & 42.9 & \textbf{50.0} & \textbf{43.5} & \textbf{51.5} & \textbf{44.6} \\
    LRP-QViT (ours) & \checkmark & \checkmark & MP6/MP6 &\textbf{45.6} & \textbf{41.3} & \textbf{48.1} & \textbf{43.0} & 49.9 & \textbf{43.5} & 51.4&\textbf{44.6}  \\
      \bottomrule
  \end{tabular}}
  \caption{Quantization results on object detection and instance segmentation on COCO dataset. Here, $AP^{box}$ is the box average precision for object detection, and $AP^{mask}$ is the mask average precision for instance segmentation.
  CRL-QViT utilizes Clipped Reparameterization for LayerNorm, while LRP-QViT incorporates both CRL and Layer-wise Relevance Propagation.
  "No Hyper-parameters" is abbreviated as "No HP", "No Reconstruction" as "No REC", and "Prec. (W/A)" indicates the quantization bit-precision for weights and activations as "W" and "A" bits, respectively. 'MP' represents mixed precision. Bold data indicate best performance.
  }
  \label{tab:ResultsCOCO}
\end{table*}

\begin{table}
  \small
  \centering
  \scalebox{0.9}{
  \begin{tabular}{lccr}
    \toprule
    Model&Method&Precision(W/A) &Top-1\%\\
    \midrule
    &Full-Precision&32/32&79.85\\
    &Layer-wise Quant~\cite{li2023repq}&4/4&33.17\\
&CW Quant~\cite{li2023repq}&4/4&70.28\\
DeiT-S&Scale Reparm~\cite{li2023repq}&4/4&69.03\\
&Clipped CW Quant (ours)&4/4&70.78\\
&Clipped CW Quant (ours)&MP4/MP4&\textbf{72.43}\\
    \midrule
    &Full-Precision&32/32&79.85\\
    &Layer-wise Quant~\cite{li2023repq}&4/4&57.63\\
&CW Quant~\cite{li2023repq}&4/4&80.52\\
Swin-S&Scale Reparm~\cite{li2023repq}&4/4&79.45\\
&Clipped CW Quant (ours)&4/4&80.55\\
&Clipped CW Quant (ours)&MP4/MP4&\textbf{81.37}\\
    \bottomrule
  \end{tabular}}
  \caption{Ablation studies of different quantizers for post-LayerNorm activation. Here, CW denotes channel-wise quantization and W/A represents weights and activation bit allocation, and MP denotes mixed precision bit allocation.}
  \label{tab:Ablation_Study_LN}
\end{table}

\begin{table}
  \small
  \centering
  \begin{tabular}{lcccr}
    \toprule
    Model&Prec. (W/A)&Top-1\%&\\
    \midrule
    &32/32&79.85 \\
    &4/4&70.78 \\
    DeiT-S&$B_1$: 5/5, $B_{2-12}$: 4/4&74.35 \\
    &$B_{1-2}$: 5/5, $B_{3-12}$: 4/4&75.66 \\
    &$B_1$: 5/5, $B_{2-12}$: LRP & 72.43 \\
    &$B_{1-2}$: 5/5, $B_{3-12}$: LRP&\textbf{72.82} \\
        \midrule
    &32/32&83.23 \\
    &4/4&80.55 \\
    &$B_1$: 5/5, $B_{2-12}$: 4/4&82.03 \\
    Swin-S&$B_{1-2}$: 5/5, $B_{3-12}$: 4/4&82.37 \\
    &$B_1$: 5/5, $B_{2-12}$: LRP & 81.13 \\
    &$B_{1-2}$: 5/5, $B_{3-12}$: LRP&\textbf{81.37} \\
    \bottomrule
  \end{tabular}
  \caption{Ablation study on mixed-precision bit allocation scheme. Here $B_{1-2}$ represents the Transformer blocks 1 and 2.}
  \label{tab:Ablation_Study_MixedPrecision}
\end{table}

\begin{table}
  \small
  \centering
  \begin{tabular}{lcccr}
    \toprule
    Model&Method&Top-1&Calib Data& Min.\\
    \midrule
    &Full-Precision&79.85&-&- \\
    &FQ-ViT~\cite{lin2022fq_vit}&0.10&1000&0.5 \\
    &PTQ4ViT~\cite{yuan2022ptq4vit}&34.08&32&3.2\\
   DeiT-S&RepQ-ViT~\cite{li2023repq}&69.03&32&1.3\\
&CRL-QViT(ours)&70.78&32&1.4\\
&LRP-QViT(ours)&\textbf{72.43}&32&1.4\\
    \midrule
    &Full-Precision&79.85&-&- \\
    &FQ-ViT~\cite{lin2022fq_vit}&0.10&1000&1.1 \\
    &PTQ4ViT~\cite{yuan2022ptq4vit}&76.09&32&7.7\\
   Swin-S&RepQ-ViT~\cite{li2023repq}&79.45&32&2.9\\
&CRL-QViT(ours)&80.55&32&3.0\\
&LRP-QViT(ours)&\textbf{81.37}&32&3.0\\
    \bottomrule
  \end{tabular}
  \caption{Comparison of the data quantity and time consumption (in minutes) during the quantization calibration.}
  \label{tab:Efficiency_analysis}
\end{table}

\subsection{Ablation Study}


To verify the effectiveness of our proposed framework, we conducted two ablation studies: clipped channel-wise quantization for post-LayerNorm, presented in Table~\ref{tab:Ablation_Study_LN}, and mixed-precision bit allocation for post-training quantization, detailed in Table~\ref{tab:Ablation_Study_MixedPrecision}.

Table~\ref{tab:Ablation_Study_LN} presents the effects in accuracy of DeiT-S and Swin-S models with various quantizer schemes for post-LayerNorm activations. For both models, direct layer-wise quantization resulted in severe performance degradation, achieving 33.17\% and 57.63\%, due to its inability to represent data distribution adequately. Applying channel-wise quantization resolved these issues, resulting in accuracy improvements to 70.28\% and 80.52\%. The scale reparameterization method, using channel-wise quantization during the quantization process and converting the learned parameters to layer-wise quantization for efficient inference, came with a drop in accuracy of about 1.25\% and 1.07\%. The clipped channel-wise method, maintained and improved the accuracy of the model by removing outliers from the post-LayerNorm parameters. This approach improved performance by 0.5\% in DeiT-S with respect to channel-wise quantization. Furthermore, adding the mixed-precision strategy further enhanced the performance of both networks by 1.7\% and 0.8\%.

Table~\ref{tab:Ablation_Study_MixedPrecision} showcases the outcomes of the classification task under various bit configurations for the quantization process. When applying 4-bit quantization to DeiT-S, enhancing the bit allocation of all layers in Block 1 ($B_1$) to 5 bits while maintaining other layers in Block 2 to 12 ($B_{2-12}$) at 4 bits raises performance from 70.78\% to 74.35\%. Similarly, allocating 5 bits to all layers in Blocks 1 and 2 ($B_{1-2}$), while keeping other blocks at 4-bits, further improves performance from 70.78\% to 75.66\%. 

The crucial importance of the first two layers in the transformer block is evident, as their output influences all subsequent layers. Therefore, in this study, we assign higher bit precision to these layers, necessitating a reduction in bit allocation from other blocks to maintain the model size at 4 bits average quantization. We achieve this by reducing the bits of the same layer but from another block, guided by the layer importance scores. Two mixed precision cases are analyzed. First, we allocate 5 bits for all layers in Block 1 and provide 3 bits to one layer 
from Blocks 2 to 12, resulting in a classification accuracy of 72.43\%. Secondly, for further enhancement, we assign 5 bits to the first two blocks and reduce bit allocation from 2 layers
from Blocks 3 to 12, resulting in 72.82\% accuracy. Our observations suggest that simply providing higher bits to the first two blocks of the transformer significantly improves performance, and the LRP method effectively identifies unimportant layers, allowing us to reduce bits and maintain the model size. The same approach is applied in the Swin-S ablation study. All results mentioned in Table~\ref{tab:ResultsImageNet} follow the second case for mixed precision, where the first two layers are given high bit precision, and the LRP method is employed to trim down the bit allocation for other less important layers

\subsection{Efficiency Analysis}
We assess the effectiveness of various methods in terms of data requirements and time consumption for quantization calibration, as detailed in Table~\ref{tab:Efficiency_analysis}. The calibration time is measured using a single 3090 GPU. FQViT, being reconstruction-free, exhibits rapid performance. However, even with 1000 calibration images, its 4-bit quantization performance is only 0.10\%. On the other hand, PTQ4ViT involves a reconstruction process, resulting in significantly higher calibration time requirements. Our LRP-QViT, requiring only 32 samples (same as RepQ-ViT), achieves a comparable calibration time, yet surpasses the performance by a substantial margin.

\section{Conclusions}
In this paper, we present a novel approach, LRP-QViT, for post-training mixed-bit quantization of vision transformers. LRP-QViT learns the layer importance score by back-propagating relevance and gradient, and based on the importance score, optimal bits are selected for mixed-precision quantization. 
Additionally, we introduce clipped channel-wise quantization for post-LayerNorm activations, which removes outliers and prevents inter-channel variation, thereby improving the model's performance. Comprehensive experiments demonstrate that our LRP-QViT outperforms existing methods in low-bit post-training quantization.